\pdfoutput=1

\documentclass[11pt]{article}

\usepackage[final]{acl}  

\usepackage{times}
\usepackage{latexsym}
\usepackage{amsmath}
\usepackage{booktabs}
\usepackage{hyperref}
\usepackage{enumitem}
\usepackage{amsfonts} 
\usepackage[T1]{fontenc}

\usepackage[utf8]{inputenc}
\usepackage{algorithm}
\usepackage{algpseudocode}
\usepackage{array}
\usepackage{multirow}
\usepackage{microtype}

\usepackage{inconsolata}
\usepackage{graphicx}
\usepackage{todonotes}

\newcommand{\new}[1]{\textcolor{black}{#1}}
\newcommand{\model}{\mbox{\textsc{ConFactCheck}}}

\usepackage{lipsum}
\usepackage{tcolorbox}
\usepackage{listings}
\tcbuselibrary{breakable, skins}

\newtcolorbox{PromptBox}[1][]{%
  colback=blue!5,       
  colframe=blue!50!black,
  coltitle=white,
  fonttitle=\bfseries,
  enhanced,
  breakable,
  title=#1,
  width=\columnwidth,
  boxrule=0.8pt,
  arc=4pt,
  outer arc=4pt,
  left=6pt,
  right=6pt,
  top=6pt,
  bottom=6pt,
}

\title{Consistency Is the Key: Detecting Hallucinations in LLM Generated Text By Checking Inconsistencies About Key Facts}

\author{
  Raavi Gupta$^{1}$\thanks{~~Equal contribution.},
  Pranav Hari Panicker$^{2}$\footnotemark[1],
  Sumit Bhatia$^{3}$,
  Ganesh Ramakrishnan$^{2}$ \\
  $^{1}$Columbia University,
  $^{2}$IIT Bombay \\
  $^{3}$Media and Data Science Research (MDSR) Lab, Adobe \\
  \texttt{raavi.g@columbia.edu}, 
  \texttt{\{pranavhp, ganesh\}@cse.iitb.ac.in}, 
  \texttt{sumit.bhatia@adobe.com}
}

\begin{document}
\maketitle

\begin{abstract}
Large language models (LLMs), despite their remarkable text generation capabilities, often hallucinate and generate text that is factually incorrect and not grounded in real-world knowledge. 
This poses serious risks in domains like healthcare, finance, and customer support. A typical way to use LLMs is via the APIs provided by LLM vendors where there is no access to model weights or options to fine-tune the model. Existing methods to detect hallucinations in such settings where the model access is restricted or constrained by resources typically require making multiple LLM API calls, increasing latency and API cost. We introduce \model{}, an efficient hallucination detection approach that does not leverage any external knowledge base and works on the simple intuition that responses to factual probes within the generated text should be consistent within a single LLM and across different LLMs. Rigorous empirical evaluation on multiple datasets that cover both the generation of factual texts and the open generation shows that \model{} can detect hallucinated facts efficiently using fewer resources and achieves higher accuracy scores compared to existing baselines that operate under similar conditions. Our code is available \href{https://github.com/decile-team/languagemodels/tree/main/ConFactCheck}{here}.
\end{abstract}

\section{Introduction}
\label{sec:intro}
Large Language Models (LLMs) are the go-to tools for NLP applications given their excellent text generation capabilities~\citep{zhao2023survey}.
However, despite recent developments in model architecture and training, even state-of-the-art models such as GPT-4~\citep{achiam2023gpt} and PALM-540B~\citep{chowdhery2023palm} often generate text that appears plausible, but is factually incorrect or non-sensical -- a phenomenon termed \textit{hallucination}~\citep{huang2023survey}. A formal analysis by \citet{xu2024hallucination} shows that LLMs cannot learn all possible computational functions, and hence, by design, will always hallucinate, albeit to different degrees. Consequently, detecting when the LLM hallucinates is imperative to take corrective action and minimize misinformation from reaching users.

Such model hallucinations can be either \textit{intrinsic} or \textit{extrinsic}~\citep{Ji_2023}. Intrinsic hallucinations arise when model outputs contradict the input or in-context instructions and can often be detected by checking input-output consistency \citep{huang2023survey}. Extrinsic hallucinations, on the other hand, occur when the model output is factually incorrect and is not grounded on the pre-training data~\citep{huang2023survey}. Given the volume of pre-training data and that it is typically inaccessible by the users, extrinsic hallucinations pose a greater challenge due to their unverifiable nature~\citep{Ji_2023}.  

\begin{figure}
    \centering
    \includegraphics[width=0.9\linewidth]{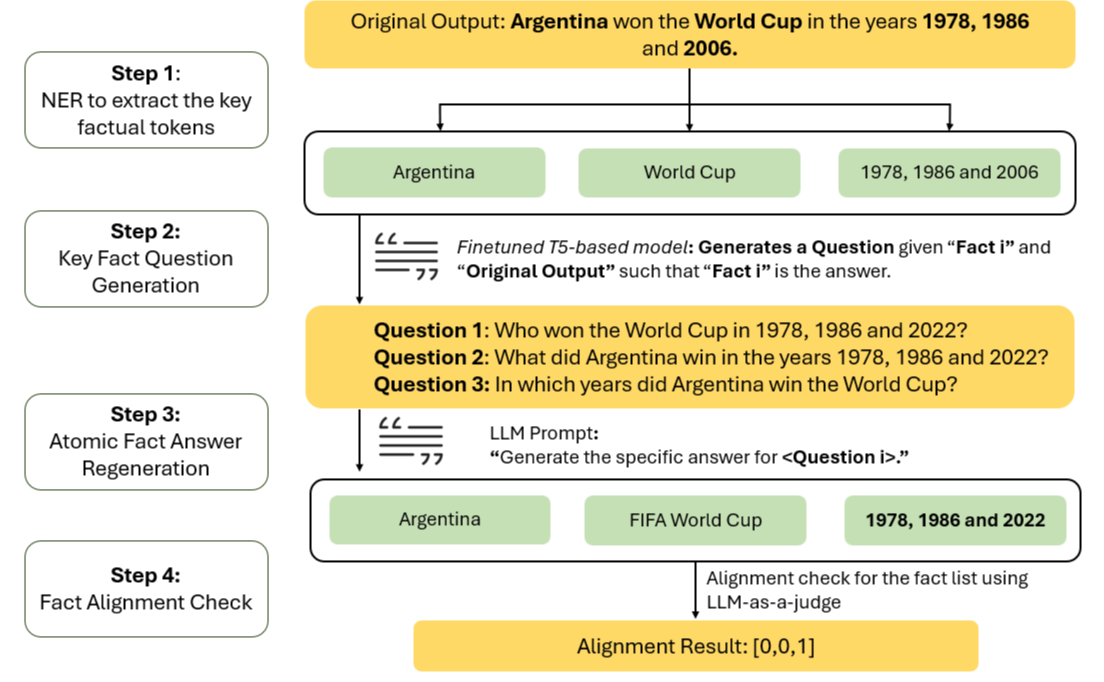}
    \caption{Key fact-based hallucination detection through the Fact Alignment check of our \model{} pipeline. Each fact is used to generate a question, and the fact is regenerated by prompting the question to the LLM. The regenerated facts are compared with the original extracted key facts to check for their consistency.}
    \label{fig:detection-pipeline-with-example}
\end{figure}

Hallucinations in LLMs are typically addressed by either \textit{(i)} improving factual accuracy via training or fine-tuning~\citep{tian2023fine, azaria2023internal, chuang2023dola}, or \textit{(ii)} verifying model outputs using external knowledge sources~\citep{cheng2024small}. 
However, in many practical cases, end-users or developers lack access to model weights or external verification sources. Recent approaches circumvent this by repeatedly querying the LLM~\citep{manakul2023selfcheckgpt, zhang2023sac3, liu2021token} to thoroughly verify responses or sample large number of outputs to estimate output probability distributions, leading to significantly increased cost and latency. To address these limitations, we propose \model, a lightweight method for hallucination detection that relies solely on the LLM’s internal knowledge. \model\ is based on a simple idea: an LLM’s understanding of a topic can be evaluated by asking related questions and measuring consistency. This recursive probing strategy has also been used in testing question-answering systems~\citep{qatest}. \new{As illustrated in Figure~\ref{fig:detection-pipeline-with-example}, \model\ identifies key entities/tags (using NER/POS tagging) in the generated output and then formulates contextually relevant questions around these entities. We term these entities/tags as `key facts', as these contain essential factual information in sentences.} The LLM's answers to these questions are checked for consistency with the original response, with high consistency indicating that the output is grounded in the model's pre-training data (reflective of the world knowledge).   

We evaluate \model \space on four different datasets spanning question-answering (NQ\_Open~\cite{nqopen}, HotpotQA~\cite{yang-etal-2018-hotpotqa}, WebQA~\cite{berant-etal-2013-semantic}) and open-ended generation tasks where inputs to the LLM lack any additional context (WikiBio~\cite{manakul2023selfcheckgpt}). \model\ outperforms recent state-of-the-art self-check or self-consistency-based baselines~\citep{manakul2023selfcheckgpt, zhang2023sac3, liu2021token} along with baselines relying on the internal states of models \citep{DBLP:conf/iclr/0026L0GWTFY24} for LLMs of different model families. \model \space achieves this outperformance while being significantly faster and requiring a lower number of LLM calls ({\em c.f.}, Table~\ref{tab:efficiency-table}). We also report the results of various ablation studies guiding our design choices and conclude by discussing the strengths and limitations of \model.

\section{Related Work}
LLMs are inherently prone to hallucinations~\cite{xu2024hallucination, Ji_2023}, a phenomenon also observed in visual and multi-modal models~\cite{bai2024hallucination, liu2024survey}. This has led to extensive research on hallucination detection and mitigation~\cite{huang2023survey, zhang2023sirens, tonmoy2024comprehensive}. Existing methods fall broadly into two categories: \textit{self-checking, prompt-based} approaches and those that require access to model weights or external knowledge sources.

\noindent
\textbf{Methods Requiring Access to Model Weights and External Sources:} \citet{tian2023fine} demonstrate that fine-tuning with factuality preferences improves output correctness. \citet{azaria-mitchell-2023-internal} use internal LLM activations passed through a classifier to estimate truthfulness. INSIDE~\citep{DBLP:conf/iclr/0026L0GWTFY24} uses internal sentence embeddings and analyzes their covariance eigenvalues to detect hallucinations. Various decoding strategies~\cite{chuang2023dola, shi2023trusting} have also been developed that utilize token probabilites at various layers to detect and mitigate hallucinations. Some approaches such as HaluAgent \citep{cheng2024small} use additional tools such as web search engines, code interpreters etc for text and code-based detection of hallucinations respectively.

\noindent
\textbf{Self-Checking and Prompt-Based Methods: }
    \citet{zhang2023sac3} propose Semantic-Aware Cross-Check Consistency (SAC$^3$), a sampling-based method that checks for self-consistency across multiple generations. Similarly, SelfCheckGPT~\citep{manakul2023selfcheckgpt} samples diverse outputs and scores their similarity to the original to estimate confidence. InterrogateLLM \citep{yehuda-etal-2024-interrogatellm}, focuses on regenerating the original query for a generated answer by reversing few-shot QA pairs to few-shot AQ pairs to self-check for model confidence during regeneration. These self-refining approaches often rely on the target LMs themselves, which is also demonstrated in Self-Refine \cite{madaan2024self}, an iterative mitigation-based approach for hallucinations. \citet{mundler2023self} explore self-contradictions using two LLMs -- one for generation and one for contradiction analysis. TRUE~\citep{honovich2022true} evaluates factual consistency using a range of metrics (n-gram, NLI, model-based) on the FEVER dataset \cite{thorne2018fever}. \citet{liu2021token} propose a reference-free, token-level method for detecting hallucinations and also present the Hallucination Detection dataset (HaDes), with raw web text being perturbed and then annotated by humans to design it for hallucination detection as a classification task. \citet{cohen-etal-2023-lm} present a cross-checking prompt-based method with 2 LLMs in a dialogue setting for evaluating hallucinations. \citet{yang-etal-2023-new-benchmark} employ a reverse validation method for self-checking via "databases" (i.e  the same LLMs), by prompting specific fact-based information to the models. FactScore \citep{min-etal-2023-factscore} breaks outputs into atomic facts, and verifies them using reliable \textit{external} knowledge sources. We also utilize the notion of atomic facts in \model\ , however, instead of leveraging external sources, we check for consistency in LLM outputs about the atomic facts.

\section{The \model{} Approach} \label{sec: method}
\begin{figure*}[h!]
    \centering
    \includegraphics[width=0.8\linewidth]{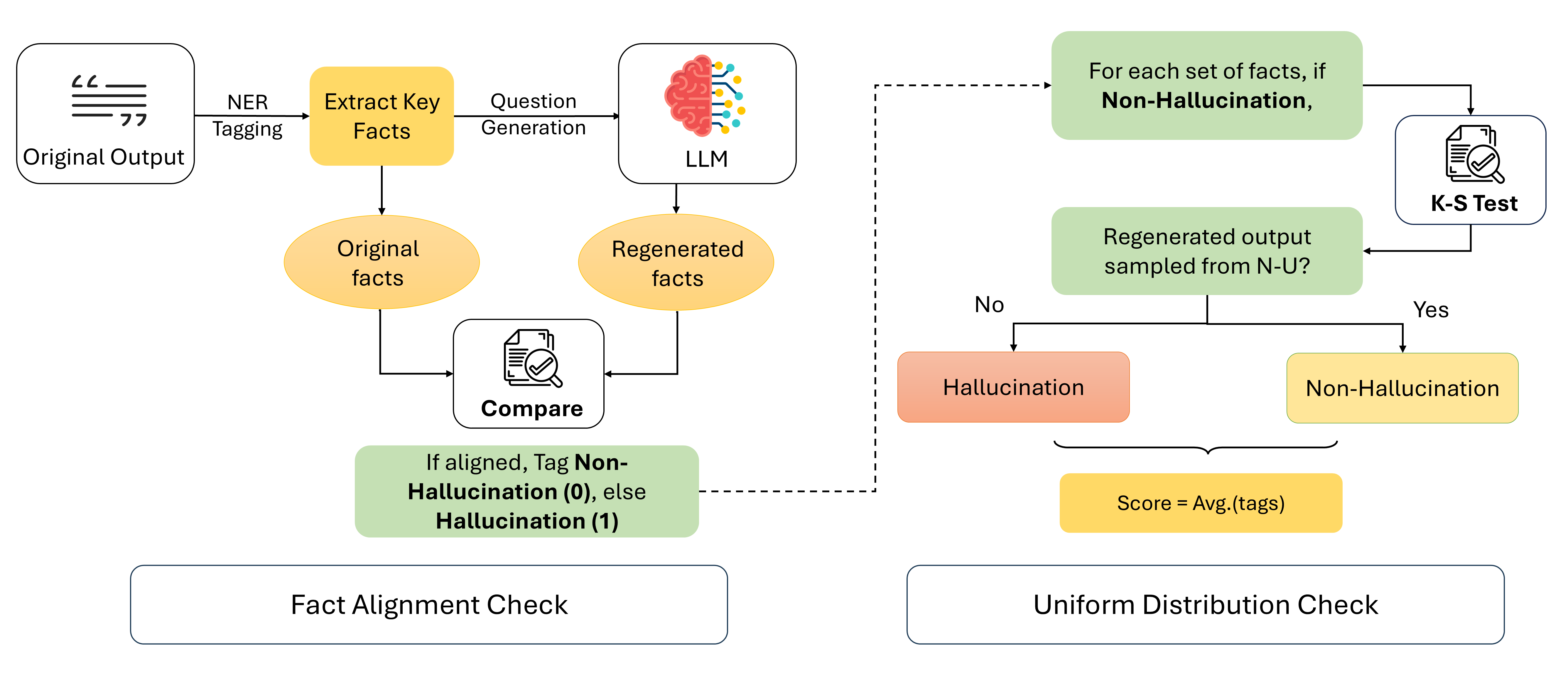}
    \caption{Pipeline of the \model \space approach, with NER tagging of outputs followed by the first comparison-based check (Fact Alignment Check) and the secondary KS test-based probability check (Uniform Distribution Check) for rechecking the classfied non-hallucinations, result in the final tagging of hallucinations. }
    \label{fig:detection-pipeline}
\end{figure*}

Figure~\ref{fig:detection-pipeline} summarizes our proposed hallucination detection approach comprising of two main steps -- \textit{(i)} a \textit{fact alignment} check where key facts in the output are compared with facts obtained by targeted probing of the LLM; and \textit{(ii)} a \textit{uniform distribution check} that filters out the low confidence predictions. We now describe the overall pipeline in detail.

\subsection{Fact Alignment Check}
\noindent
\textbf{Extracting Key Facts:} To check whether a piece of text, $\mathcal{A}$,  generated by an LLM $\mathcal{M}$ is hallucinated, we start with the assumption that the generated text is correct. 
We then generate questions targeting each key fact in $\mathcal{A}$, such that they can be answered solely using the content of $\mathcal{A}$.
Subsequently, we employ the LLM to answer the questions and see if the answers match the information in $\mathcal{A}$, a mismatch indicating hallucinations. The initial step is to identify the factual components within a sentence. According to \citet{kai2024sh2}, factual information in a sentence is typically conveyed through specific parts of speech, {\em viz.}, nouns, pronouns, cardinal numbers, and adjectives. We highlight tags with such information as key facts that are to be extracted. \citet{min-etal-2023-factscore} use a similar concept, where they classify short sentences in text (obtained by InstructGPT generation and human annotation) as atomic facts. However, the key facts we discuss are extracted NER/POS tags containing factual information, and hence are different. Key facts can be extracted by performing part-of-speech (POS) tagging or Named Entity Recognition (NER) on the sentence. Given an LLM output $\mathcal{A}$, we perform coreferencing and decompose $\mathcal{A}$ into sentences $S_1, S_2, \ldots, S_N$, where $N$ is the total number of sentences, such that $\mathcal{A} = \{S_1, S_2 \ldots, S_N\}$. Each sentence is tagged to extract key facts $a_{ij}$, where $i \in \{1, \ldots, N\}$, and $j$ depends on the number of tagged entities in a sentence. The tagging can be either POS-based or NER-based, as discussed in Section \ref{subsubsec: tagging}. For example, given the original sentence ``{\em Argentina won the World Cup in the years, 1978, 1986 and 2006.}'', in Figure \ref{fig:detection-pipeline-with-example}, the key facts consist of {\em $a = [ a_{11} = \text{Argentina}, a_{12} = \text{World Cup}, a_{13} = \text{1978, 1986 and 2006}]$}. 

\noindent
\textbf{Targeted Question Generation:} After identifying key facts, the next step involves verifying whether each fact is hallucinated within the context of the sentence. Unlike previous methodologies that assign a hallucination score to each sentence, \model{} focuses on key facts, thereby enhancing explainability by pinpointing the exact parts of a sentence that are hallucinated and providing reasons for this determination, as detailed in Section~\ref{sec: strengths}. Specifically, for each key fact $a_{ij}$ given sentence $S_i$, a corresponding question $q_{ij}$ is generated (using a T5-based model that is specifically finetuned for this task of question regeneration), with $a_{ij}$ as the target answer and $S_i$ as the context, expressed as $q_{ij} = \mathcal{Q}(a_{ij} | S_i)$, where $\mathcal{Q}$ represents the question generation module. In Figure~\ref{fig:detection-pipeline-with-example}, each key fact provides one question $q = [q_{11} = \text{Question 1}, q_{12} = \text{Question 2}, q_{13} = \text{Question 3}]$. LLM $\mathcal{M'}$ is then used to evaluate these questions at a low temperature to ensure response consistency, as it enables the LLM to generate high-quality and deterministic outputs. Each individual key fact-based question is answered by the LLM with greater precision and therefore helps to better identify whether the fact is correct or incorrect \citep{dhuliawala-etal-2024-chain}. Note that $\mathcal{M'}$ may or may not be the same as $\mathcal{M}$, as another LLM can be used to evaluate the responses of LLM $\mathcal{M}$.

\noindent
\textbf{Consistency Checking} The responses from $\mathcal{M'}$ yield regenerated facts $f_{ij}$, which are subsequently checked for consistency with $a_{ij}$. \new{To check for the similarity between $f_{ij}$ and $a_{ij}$, we follow the LLM-as-a-judge paradigm~\citep{llmasajudge}, by querying GPT4.1-mini using few-shot prompting to assess whether each pair is aligned or not.} For instance, the set $f$ for Figure \ref{fig:detection-pipeline-with-example} being {\em $f = [f_{11} = \text{Argentina}, f_{12} = \text{FIFA World Cup}, f_{13} = \text{1978, 1986 and 2022.}]$}, and original key facts being $a$ = [$a_{11}$ = \emph{Argentina}, $a_{12}$ = \emph{World Cup}, $a_{13}$ = \emph{1978, 1986 and 2006}]. In this case, facts $f_{13}$ and $a_{13}$ \new{are non-aligned; whereas, the pairs $<f_{11}, a_{11}>$ and $<f_{12}, a_{12}>$ are aligned as per the judge's output. For each aligned and non-aligned pairs, we assign the score of 0 and 1 respectively.}  Note that since the number of extracted \textit{facts} varies based on the sentence, the number of questions generated per sentence also varies. The consistency checking step, thus, enables the decomposition of sentence-level information into discrete factual elements and leverages and operates under the assumption that the LLM’s responses will remain consistent for factual information when sampled at a low temperature.

\subsection{Uniform Distribution Check} 
After the fact-alignment step, we perform a subsequent step to check if the facts were regenerated with high confidence. The underlying intuition behind this step is that if the LLM is confident in regenerating a fact correctly, the probability distribution of the generated tokens will be skewed, with the selected tokens having significantly higher probabilities than the other possible tokens. This results in a non-uniform distribution of token probabilities. Conversely, if the LLM is uncertain, even though the generated tokens may have the highest relative probability, their values will be closer to those of alternative tokens (closer to a uniform distribution) and indicating less confidence in LLM prediction. To quantify this effect, we apply the Kolmogorov–Smirnov (K–S) test to the top five tokens associated with each regenerated fact $f_{ij}$. The test is conducted using a standard significance level of $0.05$. A $p$-value below this threshold leads to the rejection of the null hypothesis (i.e.,  the top tokens are drawn from a uniform distribution) implying that the LLM exhibits confidence in its generation. If the test indicates a non-uniform distribution, the LLM is deemed confident in regeneration, and original fact $a_{ij}$ is classified as non-hallucinated. However, if the token probabilities follow a uniform distribution, it is concluded that the particular fact is hallucinated, reflecting the LLM’s lack of confidence. The final hallucination score for a sentence $S_i$ is calculated by averaging the individual scores of $a_{ij}$ present in it to give a probability of how likely a sentence has been hallucinated.

\section{Experimental Protocol}
\label{sec:experiments}

\subsection{Task and Datasets} 
\label{sec: tasks-datasets}

We consider two common task settings -- question answering (QA) and text summarization. In the QA setting, LLMs are particularly susceptible to factual hallucinations, especially when no external context or information is provided with the input questions. The summarization task is a representative of the long-form text generation tasks where the output is not limited to be a short answer (a phrase or a sentence), and hence enables us to evaluate the ability of various methods to detect hallucinations in longer pieces of text. Further, this setting also tests the ability of the LLM to generate text that is \textit{faithful} to the input context (text to be summarized). 

We use the following datasets for evaluation, (with the validation/test sets for QA):

\noindent
\textbf{1. Natural Questions (NQ)-open}~\cite{nqopen}  is an open-domain QA benchmark derived from the Natural Questions dataset~\citep{lee-etal-2019-latent}. The validation split of this dataset consists of 3,610 open-domain question-answer pairs covering a wide range of topics..

\noindent
\textbf{2. HotpotQA}~\cite{yang-etal-2018-hotpotqa} is a QA dataset that features complex questions requiring multi-hop reasoning. 

\noindent
\textbf{3. WebQA}~\cite{berant-etal-2013-semantic} dataset is a factoid QA dataset where the questions are derived from the Freebase knowledge base.   

\noindent
\textbf{4. WikiBio}~\cite{manakul2023selfcheckgpt} is a hallucination detection dataset derived from Wikipedia biographies. It consists of $238$ randomly selected articles from among the longest $20\%$ Wikipedia articles. It also provides synthetic text generated by GPT-3 for each of the original articles, along with labels for factual correctness of the sentences. 

\subsection{Baselines}
We use following four representative self-check and self-consistency based hallucination detection methods as baselines.
\\ \noindent
\textbf{HaDes}~\citep{liu2021token} is an external reference-free method that leverages various token-level features such as POS tags, average word probability, mutual information, and TF-IDF scores to identify if a token is hallucinated or not. 

\noindent
\textbf{SelfCheckGPT}~\citep{manakul2023selfcheckgpt} is a sampling based approach built upon the intuition that for hallucinated responses, stochastically sampled responses for the same input are likely to diverge. 

\noindent
\textbf{SAC$^3$}~\citep{zhang2023sac3}, another sampling-based approach that generates responses to multiple semantically similar inputs to the original input and checks for consistency in the generated outputs.

\noindent
\textbf{INSIDE}~\citep{DBLP:conf/iclr/0026L0GWTFY24} detects hallucinations using the EigenScore metric, calculated using the eigenvalues of the covariance matrix of the responses to measure the semantic consistency/diversity in the dense embedding space of the generated outputs.

\subsection{Implementation details}
\textbf{Models Used.}
We use \texttt{LLaMA3.1-8B-Instruct} and \texttt{Qwen2.5-7B-Instruct} as the base LLMs for comparing \model{} and various baselines. For the QA task, initial responses are also generated using these base LLMs, with a temperature of 1. Further, we use different models of Phi-3 family to study how well \model{} performs with LLMs of varying scale (Section ~\ref{subsubsec:scaling}). We present ablations that guided our design choices in Sections \ref{subsubsec: decoding-schemes} and \ref{subsubsec: tagging}. We use the official implementation of HaDes\footnote{\url{https://github.com/microsoft/HaDes}} for our experiments. For SAC$^3$ \cite{zhang2023sac3}, we compute the question-level consistency SAC$^3$-Q score and employ predetermined thresholds to discern the presence of hallucinated outputs.

\begin{table*}[]
\resizebox{\textwidth}{!}{\begin{tabular}{@{}lcccccccccccc@{}}
\toprule
\multicolumn{1}{c}{Model}                                    & \multicolumn{2}{c}{NQ Open}            & \multicolumn{2}{c}{HotpotQA}         & \multicolumn{2}{c}{WebQA}            & \multicolumn{2}{c}{WikiBio}          \\ \midrule
\multicolumn{1}{r}{}                                         & \texttt{LLaMA3.1}        & \texttt{Qwen2.5}        & \texttt{LLaMA3.1}        & \texttt{Qwen2.5}        & \texttt{LLaMA3.1}        & \texttt{Qwen2.5}        & \texttt{LLaMA3.1}        & \texttt{Qwen2.5}        \\ \midrule
HaDes \cite{liu2021token}                                    & 0.54              & 0.67     & 0.68     & 0.69     & 0.46              & 0.48     & N/A              & N/A     \\
$\text{SAC}^{3}$ \cite{zhang2023sac3}                        & 0.59              & 0.71              & 0.68              & 0.59              & \underline{0.63}              & 0.55              & N/A              & N/A              \\
SelfCheck-MQAG \cite{manakul2023selfcheckgpt}                  & 0.58              & 0.75              & 0.76              & 0.78              & 0.50     & 0.62              & 0.83              & 0.83              \\
SelfCheck-Prompt \cite{manakul2023selfcheckgpt}                  & \textbf{0.76}              & \underline{0.80}              & \textbf{0.86}              & \underline{0.82}              & 0.54     & \underline{0.68}              & \textbf{0.92}              & \textbf{0.90}              \\ 
INSIDE \cite{DBLP:conf/iclr/0026L0GWTFY24}                               & \underline{0.61}     & 0.54             & 0.56              & 0.60              & 0.58              & \underline{0.68}              & N/A     & N/A              \\ \midrule
\model                              & \underline{0.73}              & \textbf{0.80}               & \underline{0.83}             & \textbf{0.84}               & \textbf{0.66}               & \textbf{0.71}              & \underline{0.86}             & \underline{0.85}                         \\\bottomrule
\end{tabular}}
\caption{AUC-PR scores for NQ Open, HotpotQA, WebQA, and WikiBio datasets. We compare ConFactCheck in the same settings as the baselines, using LLaMA3.1-8B-Inst and Qwen2.5-7B-Inst as the base models. Settings for \model \space results use beam decoding on the whole pipeline (this yields best possible scores). The best performing method in a given column is in \textbf{bold} and the second best performing model is \underline{underlined}. }
\label{tab: metrics-mistral-llama}
\end{table*}

\noindent
\textbf{Metrics for Analysis:} \space\space We consider hallucination detection as a binary classification task where the text generated by the LLM is either hallucinated or not. For QA datasets, we assign labels of 1 for hallucination and 0 for non-hallucination to the original outputs by comparing them with the golden answers in the QA datasets using GPT4.1-mini as a judge LLM. For WikiBio, each sentence-level golden label is provided in the dataset itself. We compare the baselines with our approach (see Table \ref{tab: metrics-mistral-llama}) and report the AUC-PR scores on the 3 open-domain QA datasets, as well as the WikiBio summarization dataset. Note that the SelfCheckGPT baseline is applicable on the WikiBio dataset, as the others deal with only the QA task and require questions as part of their input.

\section{Empirical Results}
\label{sec:results}

\subsection{\model\ for Hallucination Detection}
Table~\ref{tab: metrics-mistral-llama} summarizes the results of different methods for the four datasets and across two LLM backbones (\texttt{LLaMA3.1-8b} and \texttt{Qwen2.5-7B}). We observe that \model{} outperforms most baselines on the QA datasets and the two LM backbones. Only the Selfcheck-Prompt baseline outdoes our approach in few settings, and even in all such cases, \model{} is the second-best performing method (the second-best approach in each column is underlined). Selfcheck-Prompt achieves the second-best performance on three other QA settings, while SAC$^3$ and INSIDE achieve the second-best performance in 2 different settings. Thereby, \model{} demonstrates consistency by either being the best or second-best performing method in all settings. Further, only SelfCheckGPT can be used for detecting hallucinations in free-form text (WikiBio dataset), as the other baselines are designed for detecting hallucinations in QA tasks and need questions as part of their input. \model{}, on the other hand, can detect hallucinations in QA as well as free-form text settings and achieves strong performance across all settings. Such strong performance of \model{} can be attributed to the fact that it identifies the \new{key factual tokens} in the generated text and probes the LLM regarding its knowledge around these tokens. 

\subsection{Computational Efficiency of Different Methods} \label{subsubsec: Efficiency}
Recall from discussions in Section~\ref{sec:intro} that self-check or self-refinement style methods suffer from high latencies due to the need to query the LLM repeatedly to estimate the output probability distributions or for a thorough verification of the generated output. \model{}, on the other hand, identifies \new{key facts} in the generated output and generates targeted questions around these facts, thereby greatly reducing the number of LLM calls. Further, \model{} relies on lightweight comparisons and statistical operations (Section~\ref{sec: method}) to check if the answers to targeted questions align with the original output. \new{Table~\ref{tab:efficiency-table} presents the average number of LLM calls made and the average inference time for different methods. We note from the table that \model{} achieves fast inference times for both the \texttt{LLaMA3.1} and \texttt{Qwen2.5} backbones. INSIDE is slightly faster than \model{}, however our pipeline offers up to $\approx$1.4x speedup compared to SelfCheck-Prompt~\cite{manakul2023selfcheckgpt} (9.51s vs. 13.35s for \texttt{LLaMA3.1}) and $\approx$1.5x and $\approx$3x when compared to SAC$^3$ (on the 2 LLMs respectively). Note also that in the case of \model{} the number of calls being made to the LLM is equivalent to the average number of key facts extracted per input in the dataset plus one additional call to the judge-LLM for Fact Alignment. In Table~\ref{tab:efficiency-table}), we report the latency numbers for SelfCheckGPT and SAC$^3$ with 20 and 5 LLM samples per question, and INSIDE with 10 LLM samples as recommended by the respective papers.} Also note that the performance numbers for SelfCheckGPT and SAC$^3$ in Table \ref{tab: metrics-mistral-llama} are with these optimal number of LLM calls (20 and 5 respectively, while they can be lower) to exhibit their best performance with efficiency. \new{All experiments on \model \space and the baselines as reported were run using NVIDIA A6000 GPUs, using the mentioned open-source LLMs for querying and execution.}

\begin{table}[h!]
\centering
\resizebox{\columnwidth}{!}{%
\begin{tabular}{@{}lcrr@{}}
\toprule
Method & \# LLM calls/samples & LLaMA3.1 & Qwen2.5 \\
\midrule
SelfCheck-MQAG & 20 & 58.9 s & 47.94 s \\
SelfCheck-Prompt & 20 & 13.35 s & 12.16 s \\
SAC$^3$ & 5 & 15.46 s & 29.37 s \\
INSIDE & 10 & 4.89 s & 5.68 s \\
\model{} & 3.8 & 9.51 s & 9.03 s \\
\bottomrule
\end{tabular}%
}
\caption{Average inference time (in seconds) for \model \space and the baselines (which have configurable amount of LLM calls) over the samples of the NQ\_Open dataset while using \texttt{LLaMA3.1} and \texttt{Qwen2.5} models. \model{} offers significant speedups over the self-checking baselines.}
\label{tab:efficiency-table}
\end{table}

\subsection{\model with LLMs of Varying Scale}
\label{subsubsec:scaling}
We now study how the performance of \model{} varies with the scale of the underlying LLM. We use the \texttt{Phi-3-Instruct} family~\cite{abdin2024phi-} of models for this purpose and chose models of 3 sizes – 3.8B, 7B, and 13B. Table~\ref{tab:performance-phi3-models} summarizes the results for the three Phi-3 models on the three QA datasets. In addition to the AUC-PR of hallucination detection, we also report the percentage of hallucinated outputs in each setting to understand the severity of hallucinations at different model scales. We note from the table that for these datasets, there is a decent amount of hallucinated outputs, which wavers from the 3.8B to 13B models. This shows that just increasing the model size may not eliminate hallucinations. We also note that the ability of \model{} to detect hallucinations is similar and consistent across different model sizes. While the Phi3-7B slightly outperforms on NQ-open, the increasing model sizes show moderate gains for the HotpotQA and WebQA datasets.
\begin{table}[h!]
\centering
\resizebox{\columnwidth}{!}{%
\begin{tabular}{@{}l cc cc cc@{}}
\toprule
\multirow{2}{*}{\textbf{Model}} & \multicolumn{2}{c}{\textbf{NQ Open}} & \multicolumn{2}{c}{\textbf{HotpotQA}} & \multicolumn{2}{c}{\textbf{WebQA}} \\  
                                & AUC             & \%Hall.            & AUC             & \%Hall.            & AUC           & \%Hall.           \\ 
\midrule
Phi-3-4b & 0.69 & 0.65 & 0.74 & 0.69 & 0.63 & 0.49 \\
Phi-3-7b & \textbf{0.73} & 0.58 & 0.74 & 0.60 & 0.62 & 0.46 \\
Phi-3-13b & 0.71 & 0.54 & \textbf{0.76} & 0.64 & \textbf{0.65} & 0.50 \\
\bottomrule
\end{tabular}%
}
\caption{Performance of \model{} for different size models of the Phi-3 family. We report AUC-PR of hallucination detection and percentage of hallucinated outputs (Hall.) for the 3.8B, 7b, and 13B models for the three QA datasets.}
\label{tab:performance-phi3-models}
\end{table}

\subsection{Ablation Studies}
We now describe different ablation studies that guided different design choices for \model{}. We report the impact of \textit{fact-alignment} and \textit{uniform distribution check} steps in the pipeline (Section~\ref{sec: method}). We also describe the effects of different decoding strategies and methods for detecting key facts in the input.

\subsubsection{Role of Different Components in \model\ }
\label{subsubec: Improvements}
Recall that there are two main steps in \model{} – \textit{fact alignment} and \textit{uniform distribution check}. The fact alignment step attempts to regenerate the key facts in the generated output by querying the LLM with targeted questions. The regenerated facts are then compared with the original output for consistency. The subsequent uniform distribution check acts as another verification layer by relying on the model's confidence in the generation of regenerated key facts. Table~\ref{tab: metrics-improve} summarizes the hallucination detection scores achieved by just the fact-alignment step along with the improvements achieved by performing the subsequent uniform distribution check (the complete pipeline). We note from the table that the uniform distribution step plays a crucial role in the overall performance of \model{} with maximum gains of up to 18\%.

\begin{table}[h!]
\centering
\resizebox{\columnwidth}{!}{%
\begin{tabular}{@{}llccc@{}}
\toprule
\textbf{Component} & \textbf{LLM} & \textbf{NQ Open} & \textbf{HotpotQA} & \textbf{WebQA} \\
\midrule
Fact Alignment & LLaMA3.1 & 0.66 & 0.79 & 0.56 \\
+ Distribution Check & LLaMA3.1 & \textbf{0.73} & \textbf{0.83} & \textbf{0.66} \\ 
\% gain & &  11\% & 5\% & 18\%  \\
\midrule
Fact Alignment & Qwen2.5 & 0.79 & 0.82 & 0.68 \\
+ Distribution Check & Qwen2.5 & \textbf{0.8} & \textbf{0.84} & \textbf{0.71} \\
 \% gain & &  1\% & 2\% & 5\% \\
\bottomrule
\end{tabular}%
}
\caption{AUC-PR scores achieved by the two major components of \model{}. A uniform distribution check after the fact alignment step leads to significant performance gains.}
\label{tab: metrics-improve}
\end{table}

\subsubsection{Effect of Decoding Strategies} \label{subsubsec: decoding-schemes}
Regardless of how the original response, subject to hallucination assessment, was generated, we examine the variations in regenerated factual responses when decoding strategies are varied. The following decoding strategies were utilized:
\begin{itemize}
    \item \textbf{Greedy Decoding}: Greedy decoding involves selecting the token from the vocabulary V with the highest conditional probability. This suggests prioritizing key facts for which the model has the highest immediate confidence.
    \item \textbf{Beam Decoding}: Beam decoding represents an enhancement over greedy decoding. In Beam decoding, a parameter known as beam\_size determines the number of tokens with the highest conditional probabilities considered at each time step t. For our experiments, we considered the beam size to be 5.
\end{itemize}

\begin{table}[h!]
\centering
\resizebox{\columnwidth}{!}{%
\begin{tabular}{@{}lcccc@{}}
\toprule
Model & NQ Open & HotpotQA & WebQA & WikiBio \\
\midrule
LLaMA3.1 (Greedy) & 0.70 & 0.81 & 0.62 & 0.86 \\
LLaMA3.1 (Beam) & \textbf{0.73} & \textbf{0.83} & \textbf{0.66} & 0.86 \\
\midrule
Qwen2.5 (Greedy) & 0.79 & 0.82 & 0.66 & 0.85 \\
Qwen2.5 (Beam) & \textbf{0.80} & \textbf{0.84} & \textbf{0.71} & 0.85 \\
\bottomrule
\end{tabular}%
}
\caption{The AUC-PR scores of \model \space with \texttt{LLaMA3.1-8B-Inst} and \texttt{Qwen2.5-7b-Inst} models using different decoding strategies for fact regeneration on the QA datasets. Beam decoding (beam size = 5) outperforms Greedy Decoding in most of the settings.}
\label{tab: metrics-decoding}
\end{table}

Beam decoding improves the detection of hallucinations during fact regeneration compared to greedy search. This advantage likely arises because beam decoding explores multiple possible answer paths before selecting the most likely one. 
Beam decoding also implicitly mitigates hallucinations by preferring sequences with higher cumulative confidence, which are more likely to reflect consistent factual patterns across generations.
As a result, when regenerating key facts, beam decoding ensures a more informed selection of entities, and the results in Table \ref{tab: metrics-decoding} show its improvements. \citet{chen-etal-2018-stable} further corroborate this by indicating that beam decoding generally outperforms greedy decoding. By maintaining multiple candidate generations, beam decoding reduces the likelihood of factual errors, ensuring the correct regeneration of facts. However, this decoding strategy does involve a trade-off with computational efficiency compared to greedy decoding.

\subsubsection{Tagging of key-facts}
\label{subsubsec: tagging}
Identifying of key facts in the generated text is a crucial step in \model{} as they are used to probe the LLM in a targeted fashion. Hence, the choice of method used for identifying key facts in the generated text can have significant impact on the overall performance. \citet{kai2024sh2} suggests that factual information in a sentence can be identified using POS tagging, specifically 'NNP' or 'NNPS'. Building on this, we selected the tags 'NNP', 'NNPS', 'CD', and 'RB' to be considered key facts. As an alternative, we also evaluated using NER tagging and considering identified named entities as key facts. We used Stanford’s Stanza~\cite{ qi2020stanza} library for NER and POS tagging. Additionally, we also sampled random tokens from the sentence and used them as key facts, ensuring that the number of sampled tokens equaled the number of NER tags present. Table \ref{tab: metrics-tagging} summarizes the results for the three strategies and reveals that though the results are similar, NER outperforms both POS tagging and random token sampling in more settings to identify which tokens contribute to the factuality of a sentence or paragraph.

\begin{table}[h!]
\centering
\resizebox{\columnwidth}{!}{%
\begin{tabular}{@{}p{2cm}cc c c c c@{}}
\toprule
Tagging & \multicolumn{2}{c}{NQ Open} & \multicolumn{2}{c}{HotpotQA} & \multicolumn{2}{c}{WebQA} \\ \midrule
        & LLaMA3.1 & Qwen2.5 & LLaMA3.1 & Qwen2.5 & LLaMA3.1 & Qwen2.5 \\
\midrule
Random & 0.72 & 0.78 & 0.82 & 0.83 & \textbf{0.68} & 0.69 \\
POS    & 0.71 & \textbf{0.81} & 0.82 & 0.83 & 0.66 & 0.7 \\
NER    & \textbf{0.73} & 0.8 & \textbf{0.83} & \textbf{0.84} & 0.66 & \textbf{0.71} \\
\bottomrule
\end{tabular}%
}
\caption{The AUC-PR scores while using different tagging strategies on LLaMA3.1-8B-Inst and Qwen2.5-7B-Inst for identifying key facts in the sentence. NER is observed to perform slightly better in more cases over these three QA datasets.}
\label{tab: metrics-tagging}
\end{table}

\subsection{Key Strengths of \model{}}
\label{sec: strengths}
We now discuss the major strengths of \model{} which are summarized as follows.

\noindent
\textbf{Training-Free Operation:}  Our generic approach requires only the LLM-generated output \new{for fact-alignment check stage} of the pipeline and does not necessitate dataset- or task-specific training.  The number of generated questions is determined by the factual content within the generated sentence, avoiding heuristic selection. During fact regeneration, \model{} leverages the output token probabilities for the probability check, which is provided by most open-source LLMs. However, in few cases of LLMs accessed via APIs it is possible that the access to output probabilities is not available (see Limitations). Even in cases where such API-based LLMs are used for generation, their outputs can be passed through open-source LLMs to perform both checks with token probabilities.

\noindent
\textbf{Ease of Implementation:} \model{} does not require access to model weights or underlying training data.  Requiring only the model's output and the LLM used for response generation, our method can be deployed on the same device as the response generation process, whether through a web interface, API, or a locally executed model. \new{Even for the use of KS test, we require only the output token probabilities of the top-5 generations, which can be directly stored during LLM generation.}

\noindent
\textbf{Consistent Sample Scoring:} Unlike stochastic hallucination detection methods such as SelfCheckGPT \cite{manakul2023selfcheckgpt}, \model{} operates deterministically by probing factual tokens at temperature 0. While similar consistency can be achieved in other methods using fixed seeds, \model{} offers this behavior by default, resulting in stable sample scoring without additional tuning. This also modestly reduces computational overhead by avoiding multiple generations per query.

\noindent
\textbf{Interpretability:} \model{} provides key-fact-level scoring, enabling users to identify specific hallucinated facts. For instance, in the running example of Figure 1, in addition to classifying the output text as hallucinated, \model\ explicitly identifies that the \textit{fact} $a_{21} = \text{\{1978, 1986 and 2006\}}$ is hallucinated (non-aligned). Operating on fine-grained facts rather than entire sentences, our pipeline offers a greater degree of explainability than previous approaches like SAC \cite{zhang2023sac3}, clarifying the rationale behind a hallucination classification.

\section{Conclusions}
We presented \model, a novel fact-based hallucination detection pipeline evaluated it using four factuality measurement datasets and compared with multiple strong baselines. Our findings reveal that despite being less computationally expensive and not requiring any training, \model{}  performs on par with other approaches while being significantly faster. 

\section{Limitations}
Despite the high performance, ease of use, and efficiency offered by \model{}, it is not without limitations. We analyze and present representative examples of failure cases to highlight its shortcomings and possible future areas of improvement. 

\noindent
\textbf{Effect of incorrect tags on correct outputs:} \\
Consider the following example from HotpotQA: \textit{Which of the office buildings used to staff the White House used to be known as the State, War, and Navy Building?} For this question, the answer provided by an LLM is the following. \textit{The office building used to staff the White House that was once known as the State, War, and Navy Building is now known as the \textbf{Eisenhower Executive Office Building}. This building was constructed in 1952 and was named after President Dwight D. Eisenhower.}

\noindent
Although Eisenhower Executive Office Building is factually correct, our pipeline categorizes the paragraph as hallucinated. This discrepancy arises because our model identifies the fact `1952' as hallucinated because of the building's actual construction period between 1871 and 1888. This contrasts with the golden output from HotpotQA, which does not flag the answer as hallucinated (when the judge LLM is used on the original output and golden answer to get the golden label). However, due to the presence of other hallucinated facts, our pipeline assigns a hallucinated tag to the paragraph.
Similarly, while the model correctly identifies the building as the Eisenhower Executive Office Building, it erroneously states the construction year as 1952 (actual: 1871–1888). As a result, \model{} tags this factual mismatch, leading to a hallucination score for the entire paragraph.

\noindent
\textbf{Inefficiency in question generation:}\\
The  generated questions extracted key facts are done by the T5-based finetuned model. While it is efficient in generating pinpointing questions with the extracted fact as answer with original output as context, some ambigious questions such as ``Who was the building named after?'' can be generated. This ambiguity can result in inaccuracies when regenerating facts. For this, using a much larger LLM can be useful, however it would be computationally expensive and time-inefficient while not providing significant improvements.

\noindent
\textbf{Language-based limited usecases:}\\
In addition, we also note that the proposed \model{} has only been tested for English language and LLMs trained mostly on English data.  
Although the framework is theoretically language-agnostic, its reliance on NER/POS tools constrains applicability in low-resource languages lacking robust NLP pipelines. Further, the performance of \model{} depends crucially on intermediate steps requiring NER and POS tagging, which may not always be available for low-resource languages.

\noindent
\textbf{Unavailability of output token probabilities in API-based LLMs:} \\
While using \model{}, fact regeneration is performed and subsequently the output token probabilities of the regenerated facts are required for the Uniform Distribution check to gauge the LLM's confidence in generation. However, it is possible that for multiple API-based LLMs or closed source models, output generation probabilities cannot be stored or utilized. Hence, \model{} cannot be accessed by such LLMs and specifically requires open-source LLMs for the two checks in the pipeline.

\bibliographystyle{acl_natbib}

\appendix
\section{Models and Implementations} \label{sec: other models}
\subsection{SelfCheckGPT \cite{manakul2023selfcheckgpt}}
One of the first papers to counter zero-resource hallucination detection, we compare the SelfCheck-Prompt variant using \texttt{LLaMA3.1-8B-Instruct} and \texttt{Qwen2.5-7B-Instruct} which is the best performing approach in their paper. Additionally we compute the SelfCheck-MQAG scores as well (which is the QA-based variant). These are presented in Table \ref{tab: metrics-mistral-llama}. We set the number of stochastic samples to be generated as 20 (as mentioned in the original paper). The scoring method for SelfCheck-MQAG selected was Bayes with Alpha. Both $\beta_1$ and $\beta_2$ were set to 0.95.

\subsection{SAC3 \cite{zhang2023sac3}}
As discussed above, for using SAC$^3$ as one of the baselines, we evaluate it using the instruction finetuned model version of \texttt{LLaMA3.1-8B} and \texttt{Qwen2.5-7B}. We calculate the question-level consistency score (SAC$^3$-Q) which is highlighted in the original study as a score describing the cross-check consistency between 2 types of QA pairs, i) the original question and generated answer as a pair and ii) a number of semantically similar generated questions along with their answers as pairs. For feasibility in accordance with our available computational resources, we experimented with 2 generated perturbated QA pairs. This number can be increased or varied to check for different comparisons, but \citet{zhang2023sac3} suggest that using between 2 to 5 perturbed questions per data sample yields similar quantitative results. 

\subsection{HaDes \cite{liu2021token}}
HaDeS is a novel token-free hallucination detection dataset for free-form text generation. For the dataset creation, raw text from web data is perturbed with out-of-box BERT model. Human annotators are then employed to assess whether the perturbed text spans are hallucinations given the original text. The final model is a binary classifier for detecting hallucinated/non-hallucinated text.

\subsection{INSIDE} \cite{DBLP:conf/iclr/0026L0GWTFY24}
INSIDE is a hallucination detection method which deals with the interal states of LLMs during generation to detect for hallucinations in outputs. Their approach utilizes the layer of sentence embedding outputs and exploits the eigenvalues of the covariance matrix of outputs to measure consistency in the dense embedding space. The define a particular score known as EigenScore, which is the logarithmic determinant of the covariance matrix between a certain K number of outputs' sentence embeddings (to check for the consistency in the relationship of those K outputs' embeddings). Using it as a baseline, we implement it with our settings with \texttt{LLaMA3.1-8B} and \texttt{Qwen2.5-7B} as the LLMs on the 3 QA datasets and calculate the AUC-PR scores.

\section{Usage of ConFactCheck on datasets}
\subsection{Open-Domain Question Answering}
Three datasets are used for this particular task, as shown above. We use ConFactCheck on the originally generated outputs for each of the questions in the datasets, to check for whether the LLMs generating the original answers have hallucinated or not. ConFactCheck is applied on a sentence-level basis, where the outputs are split into sentences, following which key facts are extracted and ConFactCheck begins the checking mechanism. 

\subsection{Text-based Summarization}
For this particular task, we use the WikiBio dataset which contains summaries of individuals collected from Wikipedia, along with synthetic GPT3 generated summaries of the same. ConFactCheck is applied as a sentence-level detector on the respective sentences of each of the provided synthetic summaries, which have be annotated with their hallucination labels at the said sentence-level as part of the dataset. We obtain sentence level hallucination scores and compare those with the golden annotate labels per sentence, and for passage-level hallucinations, we average over the sentence-level scores to get overall scores for passages.

\section{F1-Score based Matching}
In our primary pipeline, factual alignment is determined using an LLM-as-a-judge approach. Specifically, we query OpenAI’s \texttt{GPT-4.1-mini} via the API to compare extracted and regenerated facts and assign binary alignment labels. While this method yields strong performance, it requires reliable access to the OpenAI API and incurs associated computational and cost overheads. \\
To support use cases where API access is restricted or an external LLM judge is unavailable, we also explore an alternative matching strategy based on simple lexical overlap using F1-score. In this variant, alignment between fact pairs is determined by computing the F1-score of their token overlap, and pairs exceeding a predefined threshold are marked as aligned. The table below presents the AUC-PR scores across three datasets using this heuristic method at various F1-score thresholds, where the $\mathcal{M'}$ is \texttt{LLaMA3.1-8B-Instruct} (used for the fact regeneration). For this scoring, we split the extract and regenerated facts into lists of individual words, and compute the F1-scores on these lists. Different thresholds are used (as shown in Table 7 below) to assign 0/1 labels for similar/dissimilar facts.\\
Although this approach is less semantically robust than LLM-based judgment, it offers a lightweight, fully offline alternative that still provides reasonable scores that are close to the main scores in our pipeline, especially in resource-constrained settings. \\
\begin{table}[h!]
\centering
\resizebox{\columnwidth}{!}{%
\begin{tabular}{@{}lccc@{}}
\toprule
\textbf{F1-score} & \textbf{LLaMA3-NQopen} & \textbf{LLaMA3-Hotpot} & \textbf{LLaMA3-WebQA} \\
\midrule
0.4 & 0.640 & 0.791 & 0.550 \\
0.5 & 0.648 & 0.795 & 0.556 \\
0.6 & 0.659 & 0.796 & 0.556 \\
0.7 & 0.662 & 0.798 & 0.562 \\
0.8 & 0.664 & 0.800 & 0.570 \\
\bottomrule
\end{tabular}%
}
\caption{F1-score based matching with different thresholds in fact alignment (ranging from 0.4 to 0.8)}
\label{tab:f1-matching-table}
\end{table}

\section{Prompting Format}
\begin{PromptBox}[Prompt Templates Used in the Pipeline]

\textbf{1. Fact Regeneration Prompt (Manually Constructed Chat Format):}

\textit{This prompt is used to generate fact-based questions from the given sentence. 
The prompt follows a constructed chat format, to be manually customized for the model in use (e.g., LLaMA3.1, Qwen2.5). It is used for each of the questions generated by the T5-finetuned model on the extract key facts. }

\vspace{0.5em}
i) Example format for \texttt{LLaMA3-8B-Instruct}:
\begin{lstlisting}[
    basicstyle=\ttfamily\footnotesize, % Match your existing ttfamily and use a slightly smaller font if needed
    breaklines=true, % CRUCIAL: Enables line wrapping
    columns=fixed, % Ensures fixed character width
    breakatwhitespace=false, % Can break inside long strings if necessary
    showstringspaces=false % Optional: don't show visible spaces
]
'''<|begin_of_text|><|start_header_id|>system<|end_header_id|>
You are a Question-answering assistant, only answer the question.
<|eot_id|><|start_header_id|>user<|end_header_id|>
Question: <insert question here>
<|eot_id|><|start_header_id|>assistant<|end_of_header_id|>'''
\end{lstlisting}

\vspace{1em}
\textbf{2. Fact Alignment Prompt (used with the judge LLM):}
\vspace{0.5em}

\textit{Few-Shot prompt used to check for alignment between extract and regenerated facts using LLM-as-a-judge.
This prompt is well-structured to give the judge LLM complete
understanding of how to generate the alignment output for the pairs of facts that it is applied on.}
\vspace{1em}

{\ttfamily\footnotesize
'''You are a fact comparison expert. Your task is to determine whether pairs of extracted and regenerated facts refer to the same real-world entity, concept, or meaning.

For each pair:\\
- Return `0` if the two facts \textbf{refer to the same thing}, even if the wording, specificity, or structure is different.\\
- Return `1` if the two facts \textbf{do not refer to the same thing}, or if their meanings conflict.

\textbf{Guidelines:}\\
- Minor differences in wording, grammar, or capitalization should be ignored.\\
- Partial vs full names (e.g., "Vancouver" vs "Vancouver, British Columbia") should match if they refer to the same entity.\\
- Aliases and synonyms (e.g., "Roger Pirates" vs "Roger crew") should count as a match.\\
- Abbreviations (e.g., "UCLA" vs "University of California, Los Angeles") are also matches.\\
- Return `1` only if clearly unrelated or ambiguous.

\textbf{Format:}\\
Return a Python-style list of exactly \texttt{\{n\}} binary values (0 or 1), corresponding to each fact pair in order. \\
\textbf{Do not output anything else.} If unsure, still return a complete list.

\textbf{Examples:}
\begin{itemize}
  \item "President Donald J. Trump" vs "Donald Trump" → 0
  \item "Vancouver, British Columbia" vs "Vancouver" → 0
  \item "five" vs "5 seasons" → 0
  \item "UCLA" vs "University of California, Los Angeles" → 0
  \item "Microsoft" vs "Apple" → 1
\end{itemize}

Now judge the following fact pairs: \texttt{\{pairs\}} \\
Output: '''
}

\vspace{1em}

\end{PromptBox}
\vspace{0.5em} 

\noindent\begin{minipage}{\columnwidth}
    \captionof{figure}{Prompting templates used for Fact Regeneration and Fact Alignment in the \model{} pipeline. Note that the alignment prompt uses few-shot prompting.}
    \label{fig:prompt_templates} 
\end{minipage}

\section{Annotation Performance of LLM-as-a-judge}
To demonstrate the reliability of our LLM-based judging, we conducted a small-scale human evaluation. We engaged two human annotators to label 150 samples each across the three QA datasets. When comparing these human annotations with the GPT-4o labels, we observed overlap scores (between GPT and Human annotators) ranging from 82.6\% to 93\%, indicating that the LLM is capable of reliably generating accurate labels. \\\\
Furthermore, we've calculated inter-annotator agreement metrics among the human annotators as well. The Cohen's Kappa scores range from 0.76 to 0.91, which highlights substantial agreement and further corroborates the quality of our labels. We will be adding these details in the appendix of the submitted paper.

\begin{table}[h!]
\centering
\resizebox{\columnwidth}{!}{%
\begin{tabular}{@{}llc@{}}
\toprule
\textbf{Dataset} & \textbf{Metric} & \textbf{Value} \\
\midrule
\multirow{4}{*}{NQOpen} 
 & GPT Overlap (with Annotator 1) & 84\% \\
 & GPT Overlap (with Annotator 2) & 82.60\% \\
 & Inter-Annotator Overlap & 96\% \\
 & Inter-Annotator Cohen's Kappa & 91.17\% \\
\midrule
\multirow{4}{*}{HotpotQA} 
 & GPT Overlap (with Annotator 1) & 93\% \\
 & GPT Overlap (with Annotator 2) & 91\% \\
 & Inter-Annotator Overlap & 91\% \\
 & Inter-Annotator Cohen's Kappa & 78\% \\
\midrule
\multirow{4}{*}{WebQ} 
 & GPT Overlap (with Annotator 1) & 89\% \\
 & GPT Overlap (with Annotator 2) & 84\% \\
 & Inter-Annotator Overlap & 88\% \\
 & Inter-Annotator Cohen's Kappa & 76\% \\
\bottomrule
\end{tabular}%
}
\caption{GPT-human and inter-annotator overlap scores for three QA datasets (150 samples).}
\label{tab:gpt-annotator-overlap}
\end{table}

\section{Cross-evaluation with different LLMs}
\model{} provides additional flexibility when it comes to the usage of LLMs for detection in the pipeline. The original LLM used to generate the initial output can be used for the Fact Alignment check in a self-check-based setting. However, while using \model{} on the outputs of a particular base LLM (eg. \texttt{LLaMA3.1-8b-Instruct}), we can employ usage of another LLM for cross-evaluation during fact regeneration (eg: using \texttt{Qwen2.5-7b-Instruct} on the initial LLaMA-3.1 outputs). We provide experimental results to demonstrate the efficacy of cross-evaluation while using the 2 LLMs \texttt{Qwen2.5-7b-Instruct} and \texttt{LLaMA3.1-8b-Instruct} on two of the QA datasets.

\begin{table}[h!]
\centering
\resizebox{\columnwidth}{!}{%
\begin{tabular}{@{}lcc@{}}
\toprule
\textbf{Method} & \textbf{NQOpen} & \textbf{WebQ} \\
\midrule
Qwen on LLaMA3 Outputs & 0.71 & 0.63 \\
LLaMA3 as Self-Evaluator & \textbf{0.73} & \textbf{0.66} \\
\midrule
LLaMA3 on Qwen Outputs & \textbf{0.81} & 0.70 \\
Qwen as Self-Evaluator & 0.80 & \textbf{0.71} \\
\bottomrule
\end{tabular}%
}
\caption{AUC-PR scores comparing evaluator setups on NQOpen and WebQ datasets.}
\label{tab:auc-pr-evaluator}
\end{table}

\section{Comparison of Selfcheck with varying number of samples}
The SelfCheckGPT baseline methods provide configurable flexibility in terms of the number of stochastic samples that are generated to provide their final scores. The authors suggest that the samples' count can vary from 5 samples to 20 and provide similarly comparable results. We have used 20 samples for generation using Selfcheck in the Table 1 of our paper. Here, we provide a demonstration of results on the WebQA dataset with \texttt{LLaMA3.1-8b-Instruct} as the base LLM, when samples are varied between 5 and 20 for the Selfcheck methods, and compare their AUC-PR scores and computational time metrics with each other and \model{}.  
\begin{table}[h!]
\centering
\resizebox{\columnwidth}{!}{%
\begin{tabular}{@{}lccc@{}}
\toprule
\textbf{Approach} & \textbf{Sample Size} & \textbf{Time (s)} & \textbf{AUC-PR} \\
\midrule
SelfCheck-MQAG & 5 samples & 29.15 & 0.51 \\
               & 20 samples & 61.59 & 0.50 \\
SelfCheck-Prompt & 5 samples & 8.4 & 0.54 \\
                 & 20 samples & 14.1 & 0.54 \\
ConFactCheck & 2.8 avg facts + 1 API call & 8.77 & 0.66 \\
\bottomrule
\end{tabular}%
}
\caption{Performance comparison of Selfcheck methods with 5 and 20 samples each, along with latency comparisons of these approaches with ConFactCheck.}
\label{tab:hallucination-methods}
\end{table}

\section{Judge LLM vs Human evaluation in Fact Alignment}
We evaluate the reliability of LLM-based comparison in the Fact Alignment Check of the pipeline using GPT-4.1-mini as the judge LLM. Each extracted and regenerated fact pair is scored (0 for aligned, 1 for not aligned) by the LLM. To validate its accuracy, we randomly sampled 25 instances from each of three QA datasets (75 in total) and obtained equivalent 0/1 judgments from a human evaluator for each individual facts within each of the instances. Comparing the two sets of scores revealed strong agreement, with accuracy between 89.7\%–93.2\% and Cohen’s $\kappa$ > 0.79. These results offer strong evidence of the efficacy of the LLM-based comparison in our use case.
\begin{table}[h!]
\centering
\resizebox{\columnwidth}{!}{%
\begin{tabular}{@{}lcc@{}}
\toprule
\textbf{Dataset} & \textbf{Agreement (\%)} & \textbf{Cohen's Kappa} \\
\midrule
NQ-Open   & 89.7 & 0.783 \\
HotpotQA  & 93.2 & 0.845 \\
WebQA     & 89.9 & 0.799 \\
\bottomrule
\end{tabular}%
}
\caption{Model (LLaMA-3) vs. human agreement scores on the evaluation of Fact Alignment samples across three datasets.}
\label{tab:llama3-fact-alignment}
\end{table}

\section{Step-by-Step \model{} Example}

\begin{tcolorbox}[
  colback=blue!5!white, 
  colframe=blue!75!black, 
  title=Example: Question and Answer Processing Step-by-Step, 
  sharp corners, 
  boxrule=0.8pt, 
  width=\columnwidth,
  breakable
]

\textbf{Input:}
\begin{lstlisting}[
    basicstyle=\ttfamily\footnotesize,
    breaklines=true,
    columns=fixed,
    showstringspaces=false
]
Question: Who won the FIFA World Cup in 2022?
Answer: The FIFA World Cup in 2022 was won by Argentina.
\end{lstlisting}

\bigskip
\textbf{Step 1: Extract sentences from the original answer}
\begin{itemize}[leftmargin=*]
  \item The sentence splitter extracts:
  \texttt{"The FIFA World Cup in 2022 was won by Argentina."}
\end{itemize}

\textbf{Step 2: Extract Key facts using NER}
\begin{itemize}[leftmargin=*]
  \item Named entities detected: \texttt{``FIFA World Cup''}, \texttt{``Argentina''}, \texttt{``2022''}.
  \item Generated questions using T5-finetuned model for each key fact:
    \begin{itemize}
        \item \texttt{\textbf{FIFA World Cup}} $\to$ Q1: Which tournament did Argentina win in 2022?
        \item \texttt{\textbf{Argentina}} $\to$ Q2: Who won the FIFA World Cup in 2022?
        \item \texttt{\textbf{2022}} $\to$ Q3: When did Argentina win the FIFA World Cup?
    \end{itemize}
\end{itemize}

\textbf{Step 3: Generate pinpointed answers}
\begin{itemize}[leftmargin=*]
  \item Using the LM to answer the generated questions:
  \text{Answers} = [\texttt{"FIFA World Cup"}, \texttt{"Argentina"},
  \texttt{``1978, 1986 and 2022''}]
\end{itemize}

\textbf{Step 4: Compare original and regenerated answers}
\begin{itemize}[leftmargin=*]
  \item \texttt{Use Huggingface QA pipeline to extract shortened pinpointed answers from original and regenerated contexts.}
  \item Judge if answers match (0 = match, 1 = hallucination):\\
  \texttt{Initial hallucination flags = [0, 0, 1]}
\end{itemize}

\textbf{Step 5: Final hallucination check with probability}
\begin{itemize}[leftmargin=*]
  \item \texttt{Use token-level probabilities and KS-test to confirm hallucination.}
  \item \texttt{Final hallucination flags remain: [0, 1, 1]}
\end{itemize}

\end{tcolorbox}
\noindent\begin{minipage}{\columnwidth}
    \captionof{figure}{Hypothetical step-by-step example explaining the methodology of \model{}}
    \label{fig:model_step_by_step} 
\end{minipage}

\section{Pseudocode for the algorithm proposed}
The hallucination detection algorithm is designed as a two-step process applied at the sentence level for a generated answer. Given a generated answer \(\mathcal{A}\) and a model \(\mathcal{M'}\), the goal is to produce a score for each sentence indicating the likelihood of hallucination. \\
In the first step as highlighted in \textbf{Algorithm 1}, the generated answer is split into sentences, and each sentence is analyzed to extract atomic facts using Named Entity Recognition (NER). For each key fact \(a_{ij}\) in sentence \(S_i\), a corresponding question \(q_{ij}\) is generated. The model \(\mathcal{M'}\) then provides an answer \(f_{ij}\) to this question. A separate \textbf{Align} function (which uses a judge LLM for fact pair comparison) evaluates whether the fact \(a_{ij}\) is consistent with the answer \(f_{ij}\). If aligned, the fact is marked as consistent (score 0), otherwise as hallucinated (score 1). This step yields an initial binary score list for all facts. \\
In \textbf{Algorithm 2}, for each fact marked as consistent (score 0) in Step 1, we compute the logit scores of the top \(k\) tokens in the model’s answer \(f_{ij}\). These scores are converted into a probability distribution. We then perform a Kolmogorov–Smirnov (KS) test to statistically compare this empirical distribution against a uniform distribution. If the KS test yields a p-value less than a significance threshold (typically \(0.05\)), the null hypothesis — that the two distributions are the same — is rejected. This indicates that the distribution is significantly different from uniform, and the fact remains marked as consistent (score 0). However, if the p-value is greater than or equal to 0.05, the distribution is considered close to uniform, signaling high uncertainty in the model's response, and in such case the fact is reclassified as hallucinated (score 1). 
\vspace{-1em}

\vspace{1em}

\begin{algorithm}[H]
\caption{: Fact Alignment Check}
\begin{algorithmic}[1] 
    \State \textbf{Input:} Generated Answer $\mathcal{A}$, Model $\mathcal{M'}$
    \State \textbf{Output:} Initial Score List $[s_{ij}]$ for all facts $a_{ij}$
    
    \State \texttt{// Step 1: Sentence splitting and fact extraction}
    \State Perform coreference resolution on $\mathcal{A}$ and split into sentences $\{S_1, S_2, \ldots, S_N\}$
    
    \ForAll{sentence $S_i$ in $\mathcal{A}$}
        \State Extract atomic facts $\{a_{ij}\}$ from $S_i$ using NER
        
        \ForAll{fact $a_{ij}$}
            \State Generate question $q_{ij} \leftarrow Q(a_{ij} \mid S_i)$
            \State Get answer $f_{ij} \leftarrow \mathcal{M'}(q_{ij})$
            
            \If{Align($f_{ij}, a_{ij}$)}
                \State Set $s_{ij} \leftarrow 0$ \Comment{Fact is consistent}
            \Else
                \State Set $s_{ij} \leftarrow 1$ \Comment{Fact is hallucinated}
            \EndIf
        \EndFor
    \EndFor
    \State \Return $[s_{ij}]$
\end{algorithmic}
\end{algorithm}

\vspace{1em} 

\begin{algorithm}[H] 
\caption{: Uniformity Check Phase (via KS Test)}
\begin{algorithmic}[1]
    \State \textbf{Input:} Initial Score List $[s_{ij}]$, Corresponding Answer Logits $s_{ijk}$
    \State \textbf{Output:} Final Sentence Scores $[Score(S_1), \ldots, Score(S_N)]$

    \ForAll{sentence $S_i$}
        \State Initialize $Score(S_i) \leftarrow 0$
    
        \ForAll{fact $a_{ij}$ in $S_i$}
            \If{$s_{ij} == 0$}
                \State Compute normalized probabilities:
                $$p(w_{ijk}) = \frac{e^{s_{ijk}}}{\sum_{m=1}^{k} e^{s_{ijm}}}$$
                \State \texttt{// Compare with uniform distribution}
                \State Perform KS test between $p(w_{ijk})$ and uniform distribution
                
                \If{p-value $\geq 0.05$}
                    \State Set $s_{ij} \leftarrow 1$ \Comment{Mark as hallucinated}
                \EndIf
            \EndIf
            \State Add $s_{ij}$ to $Score(S_i)$
        \EndFor
        \State Normalize: $Score(S_i) \leftarrow \frac{Score(S_i)}{\text{\#facts in } S_i}$
    \EndFor
    \State \Return $[Score(S_1), \ldots, Score(S_N)]$
\end{algorithmic}
\end{algorithm}

\end{document}